\documentclass{article}

\usepackage{arxiv}

\usepackage[utf8]{inputenc} 
\usepackage[T1]{fontenc}    
\usepackage{hyperref}       
\usepackage{url}            
\usepackage{booktabs}       
\usepackage{amsfonts}       
\usepackage{nicefrac}       
\usepackage{microtype}      
\usepackage{lipsum}		
\usepackage{graphicx}
\usepackage{natbib}
\usepackage{doi}

\usepackage{amsmath,amsfonts}
\usepackage{array}
\usepackage{caption}
\usepackage{subcaption}
\usepackage{textcomp}
\usepackage{stfloats}
\usepackage{verbatim}
\usepackage{graphicx}
\usepackage{cite}
\usepackage{multicol}
\usepackage{enumitem}

\usepackage{xcolor}
\usepackage{multirow}
\usepackage{hhline}

\renewcommand{\shorttitle}{Multi-label Continual Learning for Equipment Monitoring}

\usepackage{comment}
\usepackage{placeins}
\usepackage{multirow}

\usepackage[ruled]{algorithm2e}
\makeatletter
\newcommand{\removelatexerror}{\let\@latex@error\@gobble}
\makeatother
\SetKwComment{Comment}{/* }{ */}
\usepackage{hhline}
\begin{document}

\title{A Multi-label Continual Learning Framework to Scale Deep Learning Approaches for Packaging Equipment Monitoring}

\author{ {Davide Dalle Pezze}\thanks{Corresponding author} \\
	University of Padova\\
	\texttt{davide.dallepezze@phd.unipd.it} \\
	\And
	{Denis Deronjic} \\
	University of Padova\\
	\texttt{denis.deronjic@studenti.unipd.it} \\
	\AND
	   {Chiara Masiero} \\
	Statwolf Data Science Srl\\
	\texttt{chiara.masiero@statwolf.com} \\
        \And
	{Diego Tosato} \\
	Galdi Srl\\
	\texttt{diego.tosato@galdi.it} \\
	\AND
	{Alessandro Beghi} \\
	University of Padova\\
	\texttt{alessandro.beghi@unipd.it} \\
 \And
	{Gian Antonio Susto} \\
	University of Padova\\
	\texttt{gianantonio.susto@unipd.it} \\
}



\maketitle

\begin{abstract}
Continual Learning aims to learn from a stream of tasks, being able to remember at the same time both new and old tasks.
While many approaches were proposed for single-class classification, multi-label classification in the continual scenario remains a challenging problem.
For the first time, we study multi-label classification in the Domain Incremental Learning scenario. 
Moreover, we propose an efficient approach that has a logarithmic complexity with regard to the number of tasks, and can be applied also in the Class Incremental Learning scenario.
We validate our approach on a real-world multi-label Alarm Forecasting problem from the packaging industry. For the sake of reproducibility, the dataset and the code used for the experiments are publicly available.
\end{abstract}


{\section{Introduction and main contributions}
Manufacturing machinery may now gather and communicate pertinent data regarding its condition thanks to IoT and Industry 4.0 \citep{essien2020deep,  langarica2019industrial, li2020process, maggipinto2022deep, susto2018hidden, yuan2020dynamic}. Alarm records, in particular, are frequently accessible for legacy equipment and have a far smaller memory and transmission footprint than sensor readings. Alarm analysis can therefore be a low-cost alternative or useful supplement to monitoring solutions based on raw sensor data. 
For instance, anticipating future alarms enables operators to quickly implement the best corrective measures. Alarm Forecasting (AF), plays an essential role in the safe management of process operations.
As introduced in \citep{formula}, it is frequently reasonable to rephrase AF as a multi-label classification task.

But in the industrial setting, one of the major difficulties in using Machine Learning (ML) approaches into production is that it is often difficult to adapt to process variability \citep{ma2022fault}. For example, new machines could act differently than ones that have already been deployed. Instead of learning a new model for each piece of equipment or starting from scratch every time, it would be preferable for scalability to update the current model when fresh data are available. Indeed, repeated training from scratch would require increasing amounts of memory and compute power as new units were added.
However, in the classic paradigm, model retraining using only new data leads to a sharp decrease in performance on previously learned tasks, a phenomenon known as Catastrophic Forgetting (CF) \citep{goodfellow2013empirical}. A branch of research called continuous learning (CL) tries to reduce CF and make it possible to train models using an incoming stream of training data \citep{parisi2019continual,delange2021continual}. In CL, training samples come in subsequent tasks, and the trained model can access only a single task at a time.

Proposed CL approaches for multi-label classification were proposed for the Class Incremental Learning (CIL) scenario \citep{van2019three}, where new labels appear overtime. To achieve the scalability required by a production-grade solution, we propose a CL approach for AF, framed as a Domain Incremental Learning (DIL) Multi-Label classification task. In DIL \citep{van2019three}, the labels remain the same for each task. In our setup, new machines act as a stream of new tasks. Our main contributions are the following:
\begin{itemize}
\item[(i)] We explore the use of CL techniques in a DIL scenario to scale a multi-label alarm classifier on a stream of tasks representing new pieces of equipment. To the best of our knowledge, this is the first time that DIL scenario has been investigated in a multi-label context.
\item[(ii)] We demonstrated the drawbacks of OCDM (Optimizing Class Distribution Memory), an earlier method for CIL that is transferable to DIL. We create BAT-OCDM (Balanced Among Tasks OCDM), an approach suitable for both CIL and DIL. The proposed method has a computation cost that scales logarithmically with the number of tasks $T$, whereas OCDM scales linearly with $T$.
\item[(iii)] To validate our approach, we evaluate it on a publicly available dataset of real alarm logs, coming from the packaging industry \citep{dataset}.
\end{itemize}
The rest of the paper is organized as follows. Sec. \ref{sec:related_work} provides an overview of the literature about AF and CL applied to multi-label classification. In Sec. \ref{sec:proposed_approach}, after describing a multi-label formulation for AF, we introduce a novel replay-based approach for the multi-label DIL scenario. In Sec. \ref{sec:experiments} we validate the proposed approach on a real-world AF problem coming from the Packaging Industry. Finally, Sec. \ref{sec:conclusions} provides some concluding remarks and describes envisioned future works.
}

\section{Related work}\label{sec:related_work}
The goal of alarm data analysis is to translate alarm logs into actionable insights for operators. 
Many approaches aim to detect abnormal behaviors using Anomaly Detection algorithms \citep{domingues2018comparative, yen2013beehive, du2017deeplog} or Fault Detection and Classification approaches, for example \citep{fan2020data}. 
Other approaches aim to provide useful feedback to users before abnormal behaviors occur. AF plays a fundamental role in this scenario.  
For example, in \citep{xu2019alarm} historical alarm sequences are exploited using Bayesian estimators. In \citep{zhu2016dynamic}, a probabilistic model based on an N-gram model is proposed to predict the probability of alarm occurrence, given the previous alarms. Due to the limitations of N-gram models, more advanced approaches based on neural network architectures were proposed.
In \citep{cai2019process}, alarm log information is embedded using Word2Vec and a Long short-term memory (LSTM)-based deep learning model is designed to predict the next alarm. In \citep{villalobos2020flexible}, the authors propose an approach that combines LSTM neural networks, to forecast the future measurements of various sensors, with Residual Neural Networks to predict the future occurrence of alarms based on estimated future sensor measurements.
In \citep{formula} the authors reframe AF as a multi-label classification problem and provide a more general approach that is viable in situations where raw sensor readings are not available.
\\
This work aims to adopt a CL framework to achieve a production-ready solution for AF, where to be scalable it must learn with new pieces of equipment that arrive overtime. \\

The literature often summarize the CL approaches in 3 groups, regularization-based \citep{kirkpatrick2017overcoming} \citep{zhang2022continual},  dynamic architecture methods \citep{rusu2016progressive}\citep{yoon2017lifelong} and replay-based \citep{chaudhry2019tiny} \citep{rolnick2019experience} 
CL tries to learn from a stream of tasks with non-stationary distribution. New experience is constantly being acquired overtime, while old experience is still relevant and it needs to be preserved. In the classic paradigm, model retraining using the new data leads to a sharp decrease in performance on previously learned tasks,
a phenomenon known as Catastrophic Forgetting(CF)\citep{goodfellow2013empirical}. 
The related literature suggests that the Rehearsal (also know as replay-based) approach appears to be a strong solution to CF \citep{pellegrini2020latent, buzzega2021rethinking, masarczyk2022logarithmic, kim2020imbalanced}. 
To the best of our knowledge, currently, only two main replay-based approaches designed for Multi-Label classification are proposed in the literature: 
\begin{itemize}
    \item[(i)] The first approach is Partitioning Reservoir Sampling (PRS)\citep{kim2020imbalanced}. 
The idea is that it is sufficient to allocate portion of the memory to the minority classes to retain a balanced knowledge of present and past experiences. 
    \item[(ii)] The second approach is called Optimizing Class Distribution in Memory (OCDM) \citep{liang2022optimizing}. It aims at speeding up the processing time required to select the samples, compared to PRS, obtaining a significant improvement.
OCDM is a greedy approach that selects a subset of samples such that the final distribution of the labels in memory is as close as possible to a uniform target distribution.
\end{itemize}
Both approaches are derived from the scenario of Online CL (OCL), where, in contrast to classic CL, a single batch of the task is shown at each time. Thus, in OCL, multiple epochs on the same batch are not allowed.
The PRS and OCDM were proposed for the CIL scenario\citep{van2019three} where, for each new task, new labels may appear.

In our case, we are considering a Domain Incremental Leaning (DIL) scenario\citep{van2019three} instead, where the set of output labels remains the same, but the input distribution changes based on the task. In fact, the set of alarm codes is the same for each piece of equipment, while the alarm frequency is different for the different deployments. 
In this work, we start from the OCDM to design a task-aware replay-based approach, as detailed in the next section. 

\section{Proposed approach}\label{sec:proposed_approach}
\subsection{Alarm Forecasting as a Multi-Label classification task}\label{sec:problem_statement}
To validate the proposed CL multi-label approach in the DIL scenario, we evaluate it on a real-world dataset from the dairy products packaging industry \citep{dataset}. 
In the same vein as in \citep{formula}, the goal is to produce a list of distinct alarms that are likely to occur in a future time window. 

Given the set of all alarm codes $\mathcal{A}$ and the subset of target alarm codes to be predicted $\mathcal{A}_{out} = \left\lbrace{a_1, \dots, a_L}\right\rbrace \subseteq \mathcal{A}$, the target of the multi-label classification problem is the $n$-hot encoding $y=({y}_1, \dots, {y}_L) \, \in \, \mathcal{Y}$, such that ${y}_i =1$ if alarm $a_i$ will occur in the future time window of length $D_{out}$, otherwise ${y}_i =0$. 
We represent the input data $x$ as the normalized count of distinct alarms occurred in a previous time window of fixed duration $D_{in}$. Thus, $x = \left(c_1, \dots, c_N\right) \in \mathcal{X}$, $N=|\mathcal{A}|$ and, $c_j = C_j/\sum_k C_k$ where $C_j$ is the number of alarms of type $j$ occurred in the input window.

To recap, we reframe AF as the task of estimating the following map from the input set to the label space: 
\begin{equation}\label{eq:map_h}
\begin{split}
  h \colon \mathcal{X} &\to \mathcal{Y}\\
 (c_1, \dots, c_N)  &\mapsto ({y}_1, \dots, {y}_L).
\end{split}
\end{equation}

In dealing with multi-label classification, we need to keep in mind two key challenging aspects: (i) the possibly overwhelming size of the output space, since the number of possible combinations of labels grows exponentially as the number of class labels increases \citep{zhang2013review}, and (ii) unequal label distribution in most multi-labeled datasets \citep{charte2019dealing, van2017devil,kim2020imbalanced}.
Moreover, many business-related critical alarms are rare \citep{cai2019process}.
Therefore, we use Weighted Focal Loss \citep{lin2017focal}, an idea from Object Detection and Segmentation, to train the classification model. Results from \citep{formula} indicate that this choice translates into better classification performance for low-frequency alarms.

\begin{figure}[!h]
    \centering
    \includegraphics[width=0.4\textwidth]{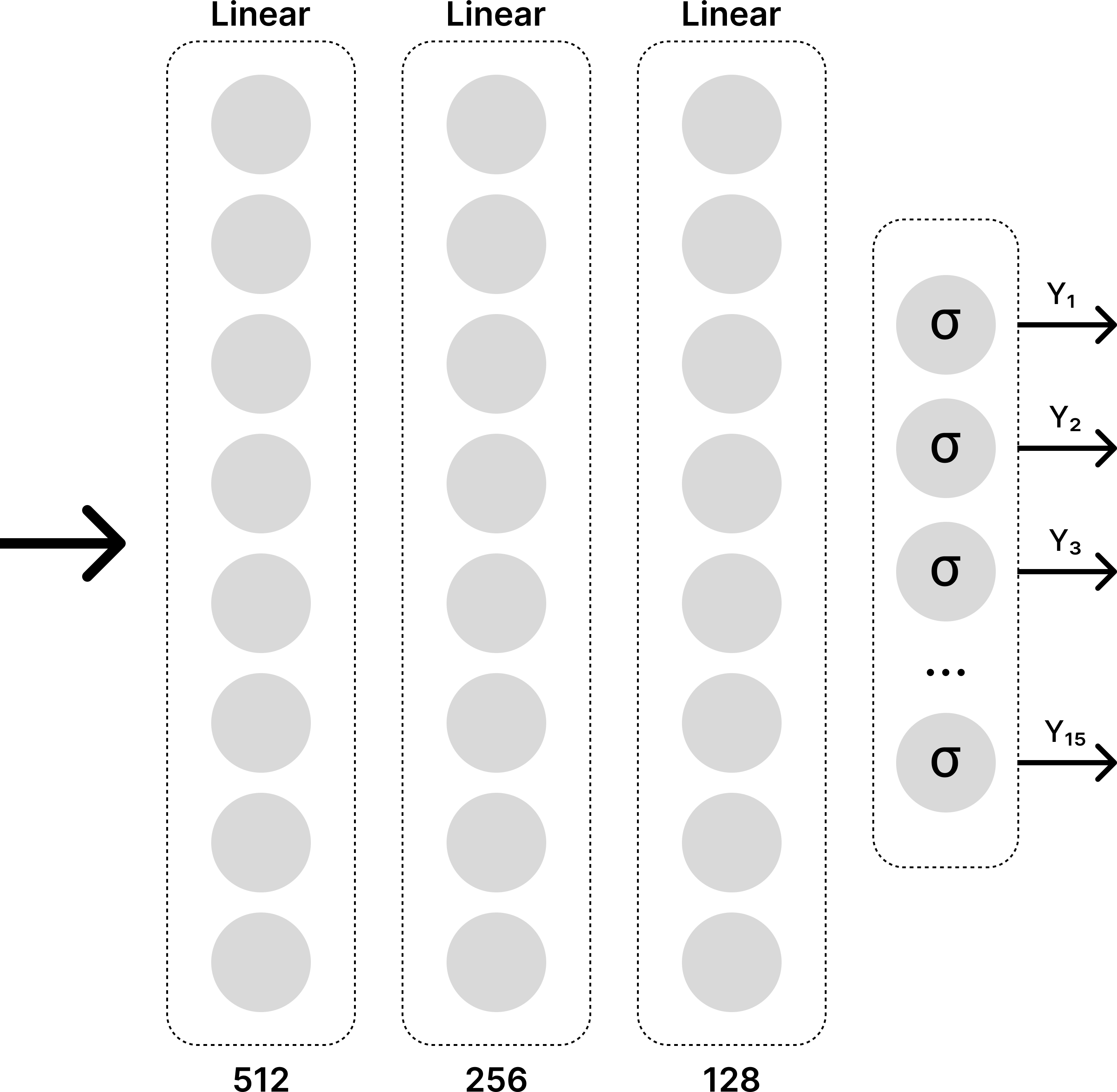}
    \caption{The multi-label architecture. Below the number of neurons in each layer. Each Linear layer has a dropout rate of 50\% as regularization and ReLU activation function except for the output layer which has independent sigmoid activation for each neuron. In this case 15 labels has to be predicted and so 15 independent sigmoid outputs.  }
    \label{fig:mlp_schema}
\end{figure}

We consider a Multi-layer Perceptron (MLP) as a model for multi-label classification. Indeed, according to \citep{formula}, approaches based on neural networks seem to achieve better performance on the AF task at hand than those based on classic ML paired with problem transformation to convert multi-label problems to single-label problems, either single-class or multi-class \citep{MADJAROV20123084, read2009classifier}.
The structure of the MLP model is detailed in Fig. \ref{fig:mlp_schema}.
Since the sigmoid activations in the output layer share the same hidden representation, this model takes into account the relations among different labels.

\subsection{Continual Learning classifier design}
\begin{figure*}[tb]
    \centering
    \includegraphics[width=0.9\textwidth]{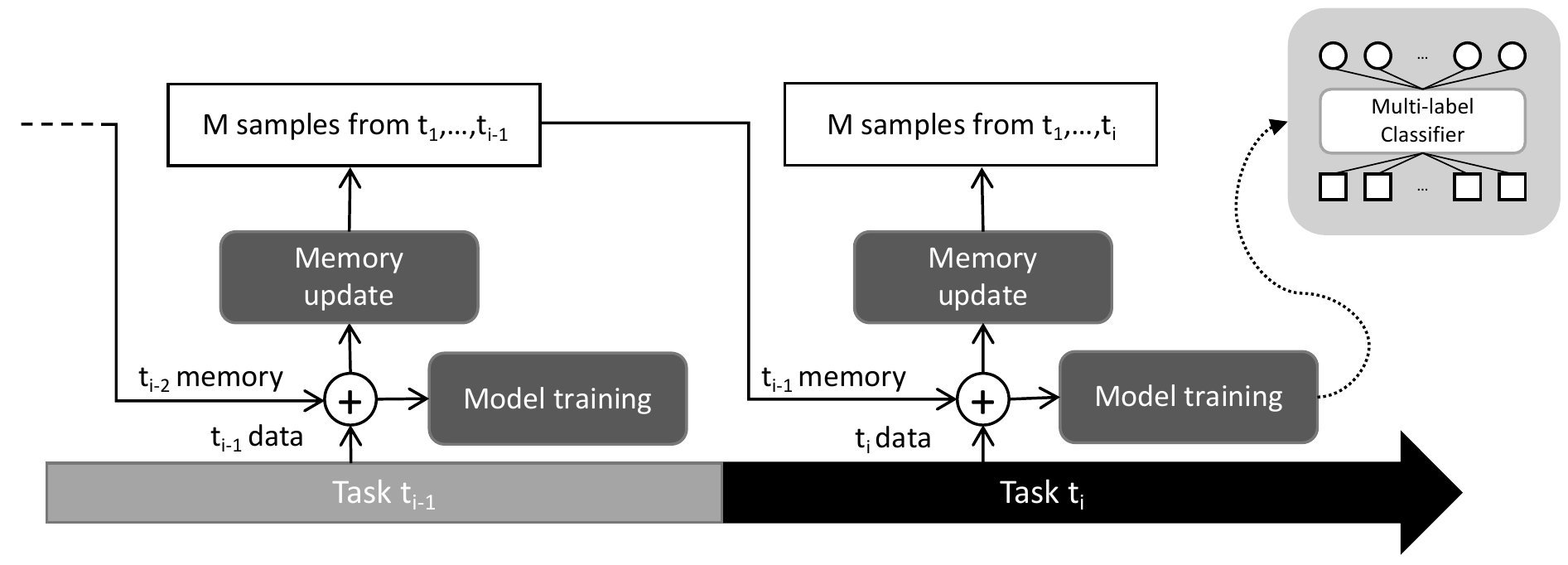}
    \caption{Rehearsal-based approach to deal with multi-label classification in a CL fashion.}
    \label{fig:rehearsal_schema}
\end{figure*}

The industrial scenario is a dynamic environment in which new machines are installed over time. CL provides tools that enable ML solutions to scale efficiently. 
In this work, we consider new equipment pieces as new tasks, setting our problem in the DIL scenario \citep{van2019three} as previously stated. This means that the set of labels in output $\mathcal{A}_{out}$ remains the same for new deployment of the target machine, while the input and output distributions change according to the current task. Indeed, even if the machines are of the same type, they may differ from the already deployed ones in terms of settings, recipes, and the behavior of human operators \citep{gentner2021dbam}. To make a Multi-Label classifier learn to perform alarm forecasting on new machines without forgetting the old ones, we use replay-based approaches from CL, according to schema depicted in Fig. \ref{fig:rehearsal_schema}.

\subsection{Optimizing Class Distribution in Memory (OCDM)}

Before introducing the novel approach used to address the Continual multi-label classification problem at hand, we recall some notions about the OCDM approach. OCDM formulates the memory update mechanism as an optimization problem. This greedy algorithm, detailed in Alg. \ref{alg:ocdm}, provides a solution with a complexity that is linear in the number of tasks.


The core of the algorithm is the procedure to update the memory when  a new batch of data $\mathcal{B}_t$ of size $b_t$ coming from the task $t \in \mathcal{T}$ arrives, as depicted in the Algorithm \ref{alg:memory_update}.
Given that the memory $\mathcal{M}$ has fixed size $M$, for each task the goal of OCDM is to select $M$ samples to save in the memory among the  $M + b_t$ available ones. 
This can be represented as an optimization problem:

\begin{equation}
\min_{\Omega} Dist(\bold{p}_{\Omega}, \bold{p}) \quad \textsc{subject to}
\quad
 \Omega \subseteq  \mathcal{M}\cup \mathcal{B}_t, \quad |\Omega| = M
\end{equation}

where \textbf{p} represents the target distribution i.e. the ideal optimal solution, while, $\textbf{p}_\Omega$ represents the distribution of the labels produced from the samples of the set $\Omega$.
The $Dist(\cdot,\cdot)$ function used to measure the difference between the two distributions is the Kullback–Leibler (KL) Distance.
The target distribution proposed in \citep{liang2022optimizing} is defined as follows:
\begin{equation}\label{eq:target_distro}
p_i =  \frac{(n_i)^\rho}{\sum_{j = 1}^{C} (n_j)^\rho}  
\end{equation}
where $n_i$ is the frequency of a class $i$ and $\rho$ is the allocation power. Using $\rho = 0$ the samples are saved in memory $\mathcal{M}$ in order to have equally distributed classes.

\begin{algorithm}[H]
\label{alg:memory_update}
\SetAlgoLined
\DontPrintSemicolon
\SetKwInOut{Input}{Input}
\caption{Memory Update (MU)}
\Input{Memory $\mathcal{M}$, b}
\BlankLine
    \Comment{Given a memory $\mathcal{M}$, it will delete $b$ elements from the memory}
    \For {$k \in [1,2,...,b]$} {
     $i = arg \min_{j \in \mathcal{M}} Dist(\bold{p}_{\mathcal{M} \setminus \{j\} }, \bold{p})$ 
     
     $\mathcal{M} \gets \mathcal{M} \setminus \{i\}$ 
}
\Return $\mathcal{M}$
\end{algorithm}

%
\begin{algorithm}[H]

\label{alg:ocdm_task}
\SetAlgoLined
\DontPrintSemicolon
\SetKwInOut{Input}{Input}
\caption{OCDM\_task\_update}
\Input{dataset $\mathcal{D}_t$ of task $t \in \mathcal{T}$ , memory $\mathcal{M}$, total size of memory M }
\BlankLine
    \For{$B_t \in \mathcal{D}_t$}
    {
        \If{$|\mathcal{M}|\leq M$}
        {
            $\text{diff} \gets |\mathcal{M}|-M$\\
            $V_t \gets$ select randomly $min(|B_t|,\text{diff})$ samples  from $B_t$\\
            $B_t \gets B_t \setminus V_t $\\
            $\mathcal{M} \gets \mathcal{M} \cup V_t$
        }
        \If{$|B_t|>0$}
        {
            $\Omega \gets \mathcal{M} \cup B_t$\\
            $\mathcal{M} \gets Memory\_Update(\Omega,|B_t|)$
        }
    }

\end{algorithm}

\begin{algorithm}[H]
\label{alg:ocdm}
\SetAlgoLined
\DontPrintSemicolon
\SetKwInOut{Input}{Input}
\caption{Optimizing Class Distribution in Memory (OCDM)}
\Input{task stream $\mathcal{T}$, total size of memory M }
\BlankLine
$\mathcal{M} \gets \{\}$ \Comment*[r]{Initialize the memory} 
\For{$t \in \mathcal{T}$} { 
    $\mathcal{D}_t \gets$ Get Dataset $\mathcal{D}_t = \{X_t, Y_t\}$ of task $t$\\
    $OCDM\_task\_update(\mathcal{D}_t,\mathcal{M},M)$
}
\end{algorithm}

\subsection{Rehearsal strategy: Balanced Among Tasks OCDM}

The OCDM approach was proposed considering the CIL scenario, where a new set of labels arrives at each new task \citep{van2019three}. As stated above, in our case, we are considering the DIL scenario, where we always have the same labels with potentially new frequencies. Since OCDM focuses only on maintaining balance among labels, it ends up ignoring the differences among tasks.

In particular, OCDM does not guarantee that all tasks seen so far will be kept in memory.
As an example, it may happen that when a new task with frequent labels arrives, then the algorithm may decide to never add any sample from this task into the memory because it would worsen the balance.

\begin{figure*}[tb]
    \centering
    \includegraphics[width=0.9\textwidth]{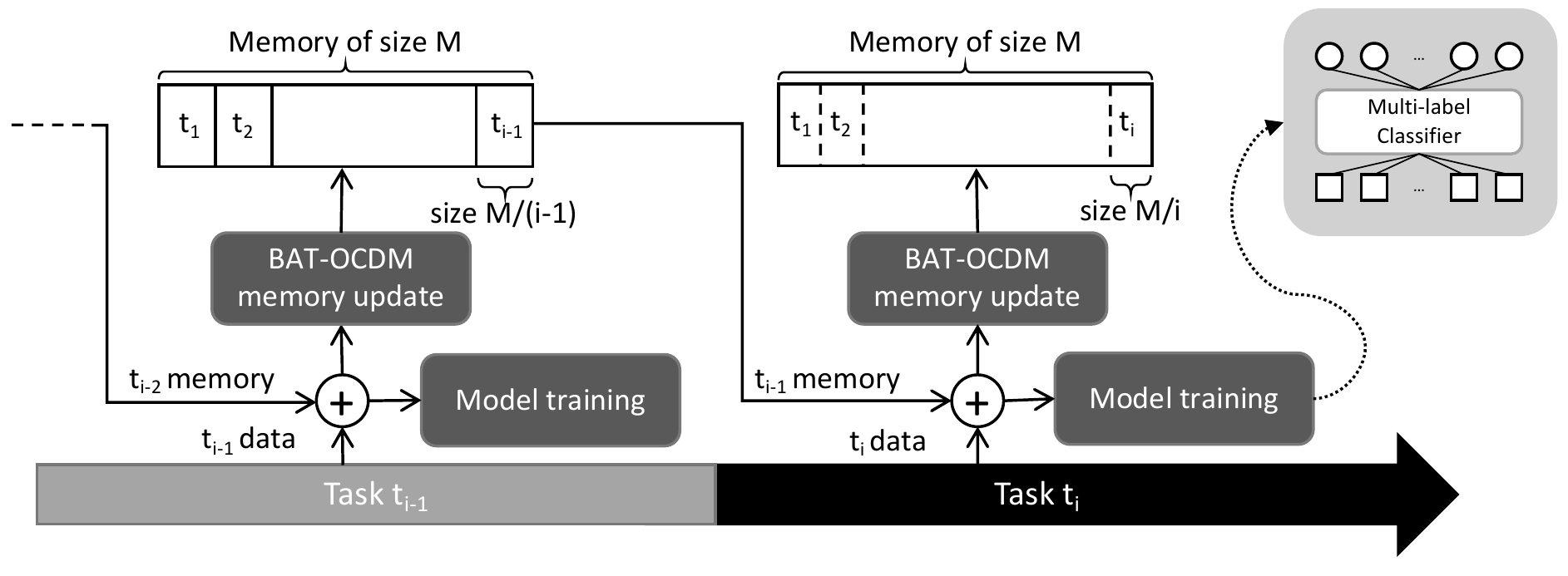}
    \caption{CL multi-label classification with BAT-OCDM strategy.}
    \label{fig:bat_ocdm_schema}
\end{figure*}

To address this limitation, we propose to use a separated memory for each task.
Through this approach, described in Alg. \ref{alg:bat_ocdm},  balance is performed on both labels and tasks, hence the name Balance Among Tasks OCDM (BAT-OCDM).
BAT-OCDM requires two steps to handle the memory:
\begin{enumerate}
    \item \textbf{Step 1}: Given the data of a new task, BAT-OCDM selects a subset of samples to be included in the memory, using the same procedure of OCDM to update the memory.
    \item \textbf{Step 2}: Since the total size of memory has to remain fixed, BAT-OCDM disregards some of the samples belonging to the memories of old tasks through the $Memory\_Update$ (MU) procedure detailed in Alg. \ref{alg:memory_update}.
\end{enumerate}
Fig. \ref{fig:bat_ocdm_schema} shows a schema of the proposed approach, which uses BAT-OCDM to train the multi-label Classifier in a CL fashion. 

\begin{algorithm}[!ht]
\SetAlgoLined
\DontPrintSemicolon
\SetKwInOut{Input}{Input}
\caption{BAT-OCDM}
\label{alg:bat_ocdm}
\Input{task stream $\mathcal{T}$, total memory size M\\
\textbf{Constraint}: $\sum_{t \in T} |\mathcal{M}_t| = M $}
\BlankLine
$N \gets 0$\\
\For{$t \in \mathcal{T}$} { 
    \Comment{We are going to assign a part of the memory to the new task and select the elements to keep in memory.}
    $\mathcal{D}_t \gets$ Get Dataset $\mathcal{D}_t = \{X_t, Y_t\}$ of task $t$\\
    $\mathcal{M}_t \gets \{\}$\\
    $N \gets N+1$\\
    $m \gets \frac{M}{N}$\\
    
    $OCDM\_task\_update(\mathcal{D}_t, \mathcal{M}_t, m)$
    
    \BlankLine
    \Comment{We update the memories of old tasks, removing some elements, to give space in memory to the new task.}
    $\text{diff} \gets \frac{M}{N-1}-\frac{M}{N}$\\
    \If{$N \geq 2$}
    {
        \For{$t_{old} \in \mathcal{T}_{1 \cdots N-1}$}
        {
            $\mathcal{M}_{t_{old}} \gets Memory\_Update(\mathcal{M}_{t_{old}}, \text{diff})$
        }
    }
    
}
\end{algorithm}

\begin{algorithm}[!ht]
\SetAlgoLined
\DontPrintSemicolon
\SetKwInOut{Input}{Input}
\caption{Dataset-based OCDM}
\label{alg:complete_ocdm}
\Input{task stream $\mathcal{T}$, total size of memory M  }
\BlankLine
$\mathcal{M} \gets \{\}$ \Comment*[r]{Initialize the memory} 
\For{$t \in \mathcal{T}$} { 
    $\mathcal{D}_t \gets$ Get Dataset $\mathcal{D}_t = \{X_t, Y_t\}$ of task $t$\\
    $\mathcal{M} \gets \mathcal{M} \cup \mathcal{D}_t$\\
    \If{$|\mathcal{M}| \geq M $}
    {
        $\Omega \gets \mathcal{D}_t \cup \mathcal{M}$\\
        $diff \gets min( |\Omega| - M, M) $\\
        $\mathcal{M} \gets Memory\_Update(\mathcal{M},diff)$
    }

}
\end{algorithm}

\subsubsection{Computational complexity}
Next we compare BAT-OCDM and OCDM in terms of computational complexity. We also consider Dataset-based OCDM, a version of OCDM where, instead of sequential batches, all the data coming from the task are used for the memory update. This should improve performance, since the original batch-based memory update policy could ignore some samples, and thus it may lead to a less uniform label distribution in the final memory. On the other hand, in this way the update can no longer be performed on-line as in the original OCDM. Tab. \ref{tab:results_performance} summarizes the computational complexity of the strategies mentioned above, while more details can be found in the supplementary material \citep{repo_bat_ocdm}.

\begin{table}
\centering
\caption{Notation used to show the complexity of the algorithms}
\begin{tabular}{|l|l|} 
\hline
Notation & Description                               \\ 
\hline
D        & size of the dataset associated to a task  \\ 
\hline
T        & number of tasks used in the experiment    \\ 
\hline
M        & size of the memory used                   \\
\hline
\end{tabular}
\end{table}
\begin{table*}[!h]
\centering
\begin{tabular}{|p{15mm}|l|l|l|}
\hline
              & Task i                                         & Total & Assuming $M=c \cdot D$  \\ \hline
OCDM          & $O(D \cdot M)$                                   &  $O(T \cdot D \cdot M)$  & $O(c \cdot T \cdot D^2)$    \\ \hline
Dataset-based OCDM & $O(D \cdot M + \frac{D^2}{2})$                  &  $O( T \cdot [ D \cdot M + \frac{D^2}{2} ])$ & O( $\frac{2c+1}{2} \cdot T \cdot D^2)$     \\ \hline
BAT-OCDM      & $O( \frac{D \cdot M}{i} + \frac{M^2}{i-1})$     &  $O( (\ln T +1) \cdot M \cdot (D+M) ) $ & $O( (c+c^2) \cdot (\ln T +1) \cdot D^2)$  \\ \hline  
\end{tabular}
\caption{Computational complexity of the proposed approaches}
\end{table*}

As for the notation, we denote the set of tasks as $\mathcal{T}=\{t_1, \dots, t_T\}$. For simplicity, we assume that each task corresponds to $D$ samples and the memory $\mathcal{M}$ has a fixed size M.

The column \textit{Task i} provides the complexity of the $i-$th memory update.
For BAT-OCDM the complexity becomes smaller as $i$ increases. In particular, the overall complexity, shown in column \textit{Total}, is logarithmic in the number of tasks $T$ for BAT-OCDM. For OCDM and Dataset-based OCDM instead, the complexity is linear in $T$.

The last column shows the complexity under the assumption that $M\leq D$, i.e. $M=c \cdot D$ where $c \in [0,1]$. In this case, the complexity for all methods is quadratic with regard to to the dataset size D. The results on computational complexity do not hold only for the DIL scenario, but also for the CIL scenario studied in the original paper proposing OCDM, showing the same speed improvement.

\section{Experiments}\label{sec:experiments}
\subsection{Metrics}
\label{sec:metrics}
{Following the convention of multi-label classification \citep{wang2016cnn} \citep{ge2018multi} we are going to use the macro f1 score to evaluate the performance of the model.}
Let $s$ be the macro $f_1$ score, i.e. the average of the $f_1$ scores for each label: $s = \sum_{i=1}^L f_1(y_i, \hat{y_i})$.
Let $s_{i,j}$ be the performance of the model on the test set of task $j$ after training the model on task $i$. To measure performance in the CL setting, we introduce the following metrics:
\begin{description}
\item[Average macro f1] 
The average macro f1 score $S_T \in [0,1]$ at task T is defined as:
\begin{equation}
    S_T = \frac{1}{T} \sum_{j=1}^T s_{T,j}
\end{equation}

\item[Average Forgetting] $F_T \in [-1,1]$, the average forgetting measure at task T, is defined as:
\begin{equation}
    F_T = \frac{1}{T-1} \sum_{j=1}^{T-1} max_{l \in \{1,\cdots,T-1\} } \frac{s_{l,j}-s_{T,j}}{s_{l,j}}.
\end{equation}
With respect to the original definition used in \citep{chaudhry2018riemannian}, we are scaling respect to the maximum macro f1 score, as done in \citep{kim2020imbalanced}; this is done to compare the forgetting among labels with very different scores. Notice that the closer the metric $F_T$ is to 1, the higher the forgetting is.
\end{description}

\subsection{Experimental setup}
We test the proposed approach on a publicly available real industrial dataset originating from the monitoring of dairy products packaging equipment \citep{dataset}. In the experiments, we consider the monitoring of 14 packaging machines deployed in different plants around the world as different tasks to learn in a CL fashion.
The splitting of data in train and test is based on time where in the test there will be only samples belonging to the future of the machine.\\
To obtain the design matrix to train the CL-based multi-label classifier, we draw inspiration from \citep{formula}, so we consider input windows having a length of $1720$ minutes, and output windows of $480$ minutes. As described in Sec. \ref{sec:problem_statement}, we represent the input windows using normalized alarm counts.

\begin{figure*}
 \begin{subfigure}[b]{0.45\textwidth}
    \centering
    \includegraphics[width={\textwidth}]{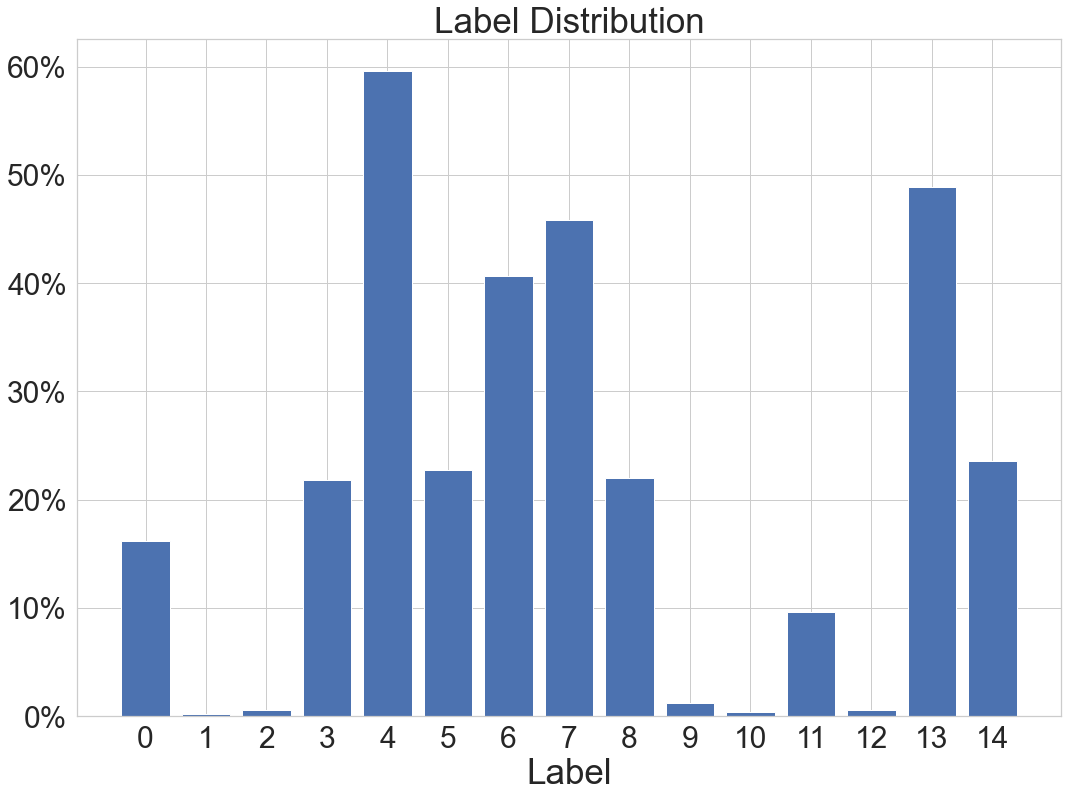}
    \caption{Label frequency}
    \label{fig:label_distro}
\end{subfigure}
\hfill
 \begin{subfigure}[b]{0.45\textwidth}
    \centering
    \includegraphics[width={\textwidth}]{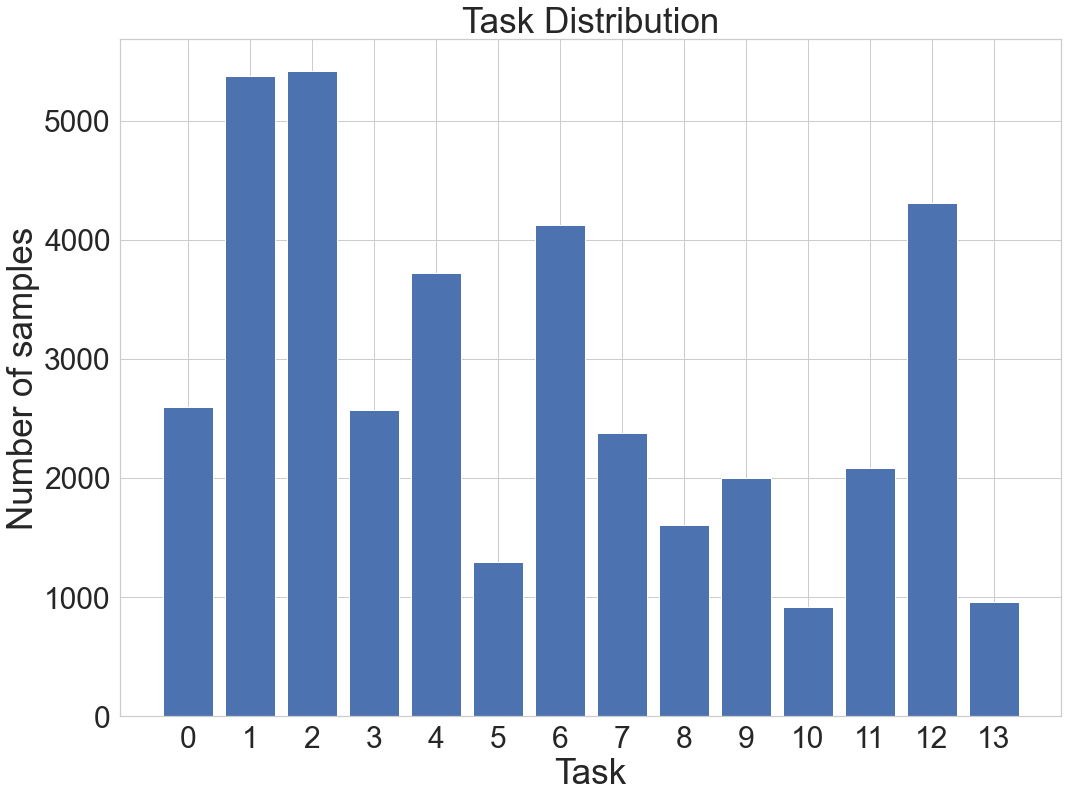}
    \caption{Number of samples per task}
    \label{fig:task_distro}
\end{subfigure}
\caption{Experimental dataset: Labels and tasks statistics.}
\label{fig:dataset_distribution_plots}
\end{figure*}

Fig. \ref{fig:dataset_distribution_plots} shows in detail the frequencies of the labels and the number of samples for each task. 
Based on label frequencies, we split them into three groups to better assess the performance: 
\begin{itemize}
    \item high-freq: labels \{4,6,7,13\}; 
    \item medium-freq: labels \{0,3,5,8,14\}; 
    \item low-freq: labels \{1,2,9,10,11,12\}.
\end{itemize}

We used grid search to select the hyper-parameters for the rehearsal approaches. The \textit{ratio of replay} is set to 0.5, i.e. at each batch 50\% of data will be from the new task and 50\% of the data will be from the old tasks.
The \textit{memory size} is set to 2000 samples which corresponds also to the 5\% of the entire set of samples.
Finally, the hyperparameter $\rho$ used in OCDM and BAT-OCDM is set to zero, as suggested in the original paper.

All the experimental settings-related details are available at \citep{repo_bat_ocdm} for reproducibility.

\subsection{Considered Continual Learning approaches}
All the memory management strategies can be schemed according to the framework shown in Fig. \ref{fig:rehearsal_schema}.
To assess BAT-OCDM, we compare it not only with OCDM (Alg. \ref{alg:ocdm}) and Dataset-based OCDM (Alg. \ref{alg:complete_ocdm}), but also with the following approaches:
\begin{itemize}
\item Finetune: At each task $t_i$, the classifier is trained from scratch considering only the data from task $t_i$. There is no countermeasure against forgetting, so we consider this approach as a lower bound.
\item Cumulative: At each task $t_i$, the classifier is trained from scratch based on all the data seen so far (i.e. from task $t_1$ to $t_i$), so there is no limit to memory size.
\item Task-based Random: At each task $t_i$, an equal number of samples is selected at random.
\item Reservoir Sampling: This stream-oriented approach from Online Learning selects samples at random at a fixed rate. Thus, it is expected to include more samples for tasks whose dataset size is bigger.
\end{itemize}

The code for the BAT-OCDM approach is shared publicly\footnote{ https://github.com/dallepezze/bat-ocdm } as well the dataset used \citep{dataset}.

\subsection{Experimental results}
We divide alarm codes into three categories based on their frequency in order to more thoroughly assess how the label frequency influences model performance. As stated, we then calculate the performance measures for low, medium, and high-frequency alerts independently. The computation time (measured in seconds) needed for the memory update for each suggested strategy is also evaluated in addition to the average macro f1 score and the forgetting measure.

\subsubsection{Performance}
The experimental results are displayed in Tab. \ref{tab:results_performance}. Among the two approaches based on the random selection of the samples, there is no significant difference. It is clear from the results that task-based Random and Reservoir Sampling appear to work well on labels with high and medium frequencies. Instead, outcomes are unsatisfactory when low-frequency labels are taken into account, because of the poor representation of these labels in the memory. 
Unbalanced label distribution is one of the difficulties with multi-label, since it makes it harder for the model to concentrate on the rare labels and results in poor end performance. The best choice of samples to keep in memory also makes the multi-label in continual learning more challenging (with respect to the classic multi-label setting) because it must come up with a good strategy to keep a portion of samples from previous tasks in memory while attempting to have a balanced label distribution that gives a more equal representation to all labels. 
 The OCDM method, in contrast, produces good performance for low frequency labels. These labels have a critical impact on the equipment as evidenced in \citep{formula}.  However, this result comes with the trade-off of less effective performance on labels with high frequency. Instead, there is no relevant performance difference for the classification of medium-frequency labels.

A possible explanation for this, as mentioned above, could be that some tasks are not stored in memory. Indeed, the algorithm focuses only on the balance among labels and not among tasks. This will be explained in more details in Sec. \ref{sec:balance_among_labels} and in Fig. \ref{fig:task_distro} with the task distribution in memory, where it can be seen the very low balance among tasks. For this reason, BAT-OCDM shows superior performance respect to OCDM for low-freq, medium-freq and high-freq labels. In particular, the drop in performance for higher frequency labels observed for OCDM is much lower. 
As for Dataset-based OCDM, even if it simultaneously uses all data from a task to find a better label distribution in memory, no significant difference appears in the performance. Instead, as anticipated by the complexity analysis, the time required to update the memory is much higher.

\begin{table}
\centering
\caption{The table contains the performance for each approach. In each cell are showed two metrics defined like in Sec. \ref{sec:metrics}. Above is the Average macro f1 $S_T$ and below the Average Forgetting $F_T$. Based on the column, these metrics are calculated on a different set of labels. \textit{Low}, \textit{Medium}, and \textit{High} are label sets grouped by the frequency of the labels, while \textit{Total} consider all the labels together.}
\label{tab:results_performance}
\begin{tabular}{|c|c|c|c|c|c|} 
\hline
Approach                                                                           & \multicolumn{1}{l|}{Total} & \multicolumn{1}{l|}{Low} & \multicolumn{1}{l|}{Medium} & \multicolumn{1}{l|}{High} & \multicolumn{1}{l|}{Time (s)}                               \\ 
\hline
\multirow{2}{*}{Finetune}                                                          & 0.19                       & 0.04                     & 0.21                        & 0.38                      & \multirow{2}{*}{-}                                          \\
                                                                                   & (0.27)                     & (0.14)                   & (0.39)                      & (0.33)                    &                                                             \\ 
\hline
\multirow{2}{*}{Cumulative}                                                        & 0.39                       & 0.17                     & 0.42                        & 0.7                       & \multirow{2}{*}{-}                                          \\
                                                                                   & -                          & -                        & -                           & -                         &                                                             \\ 
\hhline{|======|}
\multirow{2}{*}{\begin{tabular}[c]{@{}c@{}}Task-based\\Random\end{tabular}}        & 0.37                       & 0.14                     & \textbf{0.4}                & 0.69                      & \multirow{2}{*}{0.3}                                        \\
                                                                                   & (0.11)                     & (0.09)                   & (0.18)                      & (0.05)                    &                                                             \\ 
\hline
\multirow{2}{*}{\begin{tabular}[c]{@{}c@{}}Reservoir \\Sampling (RS)\end{tabular}} & 0.37                       & 0.14                     & \textbf{0.4}                & \textbf{0.69}             & \multirow{2}{*}{0.3}                                        \\
                                                                                   & (0.1)                      & (0.1)                    & (0.16)                      & \textbf{(0.04)}           &                                                             \\ 
\hline
\multirow{2}{*}{OCDM}                                                              & 0.32                       & \textbf{0.2}             & 0.32                        & 0.49                      & \multirow{2}{*}{2607.9}                                     \\
                                                                                   & (0.2)                      & \textbf{(0.04)}          & (0.31)                      & (0.31)                    &                                                             \\ 
\hline
\multirow{2}{*}{\begin{tabular}[c]{@{}c@{}}Dataset-based \\OCDM\end{tabular}}      & 0.32                       & \textbf{0.2}             & 0.33                        & 0.49                      & \multirow{2}{*}{\textcolor[rgb]{0.114,0.11,0.114}{4702.3}}  \\
                                                                                   & 0.2                        & \textbf{0.04}            & 0.31                        & 0.31                      &                                                             \\ 
\hline
\multirow{2}{*}{BAT-OCDM}                                                          & \textbf{0.38}              & \textbf{0.2}             & \textbf{0.4}                & 0.64                      & \multirow{2}{*}{\textbf{624.4}}                             \\
                                                                                   & \textbf{(0.09)}            & \textbf{(0.04)}          & \textbf{(0.15)}             & (0.085)                   &                                                             \\
\hline
\end{tabular}
\end{table}

\subsubsection{Balance among labels}
\label{sec:balance_among_labels}
\begin{figure}[tb]
 \centering
    \begin{subfigure}[b]{0.45\textwidth}
        \centering
        \includegraphics[width=\textwidth]{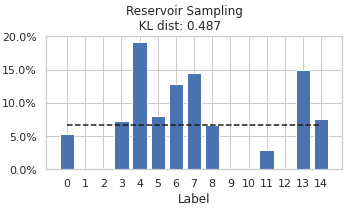}
        \caption{Label distribution for Reservoir Sampling}
        \label{fig:label_distro_rs}
    \end{subfigure}
    \hfill
    \begin{subfigure}[b]{0.45\textwidth}
        \centering
        \includegraphics[width=\textwidth]{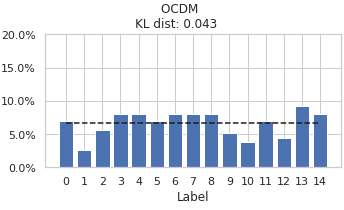}
          \caption{Label distribution for OCDM}
        \label{fig:label_distro_ocdm}
    \end{subfigure}
    \\
    \begin{subfigure}[b]{0.45\textwidth}
        \centering
    \includegraphics[width=\textwidth]{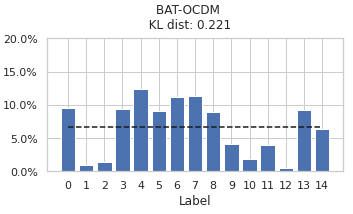}
     \caption{Label distribution for BAT-OCDM}
        \label{fig:label_distro_bat_ocdm}
    \end{subfigure}
    \caption{Final distribution of the labels of the samples in memory varying the approach. In the title for each plot is indicated the name of the approach and KL distance respect to the target uniform distribution: the larger the value, the less balanced the labels.
 }
    \label{fig:distribution_labels_memory}
\end{figure}

We can see the label distributions from the RS, OCDM, and BAT-OCDM techniques in Fig. \ref{fig:distribution_labels_memory}. Each plot also shows the KL distance to indicate how far apart the final distribution is from the desired distribution defined in Eq \eqref{eq:target_distro}. We can observe that OCDM better approximates the target distribution. As opposed to OCDM, Random Selection displays a KL distance that is roughly ten times larger. This is a result of the heavily unbalanced dataset that was employed.
When Fig. \ref{fig:label_distro} and plot Fig. \ref{fig:label_distro_rs} are compared, it can be observed that because the selection is made randomly, the memory has the same label distribution as the dataset. So the distribution in memory will be very far from the target distribution with a high KL distance. 
Regarding BAT-OCDM, it considers samples from all of the available tasks. The resulting distribution shows a KL distance five times higher than OCDM and two times lower than RS. 

\subsubsection{Balance among tasks}

\begin{figure}[tb]
 \centering
    \begin{subfigure}[b]{0.45\textwidth}
        \centering
        \includegraphics[width=\textwidth]{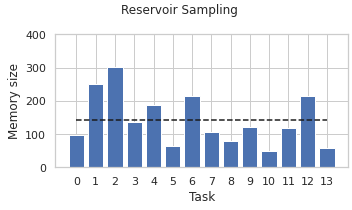}
        \caption{Task distribution for Reservoir Sampling}
        \label{fig:task_distro_rs}
    \end{subfigure}
    \hfill
    \begin{subfigure}[b]{0.45\textwidth}
        \centering
        \includegraphics[width=\textwidth]{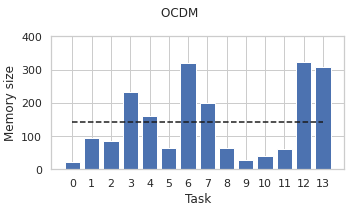}
          \caption{Task distribution for OCDM}
        \label{fig:task_distro_ocdm}
    \end{subfigure}
    \\
    \begin{subfigure}[b]{0.45\textwidth}
        \centering
    \includegraphics[width=\textwidth]{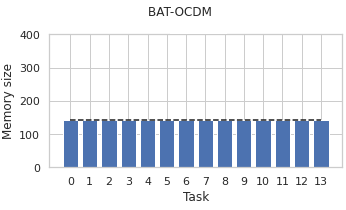}
     \caption{Task distribution for BAT-OCDM}
        \label{fig:task_distro_bat_ocdm}
    \end{subfigure}
    \caption{Final distribution of the number of samples for each task in memory varying the approach. The dashed line in each plot represents the average number of samples among all the machines. On the x-axis of the plots are represented the task IDs and on the y-axis the number of samples of each machine as absolute value.}
    \label{fig:distribution_machines_memory}
\end{figure}

The task distribution in memory achieved by the RS, OCDM, and BAT-OCDM approaches is shown in Fig. \ref{fig:distribution_machines_memory}. By design, BAT-OCDM achieves a completely uniform allocation of the tasks (Fig \ref{fig:task_distro_bat_ocdm}). Additionally, Task-based Random has the same characteristic because it maintains a separate memory for every task. For the Reservoir Sampling technique, the amount of samples in memory for each task is proportional to the dataset size of that task as shown by Figs. \ref{fig:task_distro_rs} and \ref{fig:task_distro}. This is  because the random selection occurs at a fixed rate during the stream of samples corresponding to the various tasks.
About OCDM and Dataset-based OCDM, they have a similar distribution shown in the center of Fig. \ref{fig:task_distro_ocdm}. In this case, we can observe that some tasks are almost absent in the memory, with the risk of being forgotten more quickly than the other tasks.

\subsubsection{ Computation times}
\begin{figure}[]
    \centering
    \includegraphics[width=0.49\textwidth]{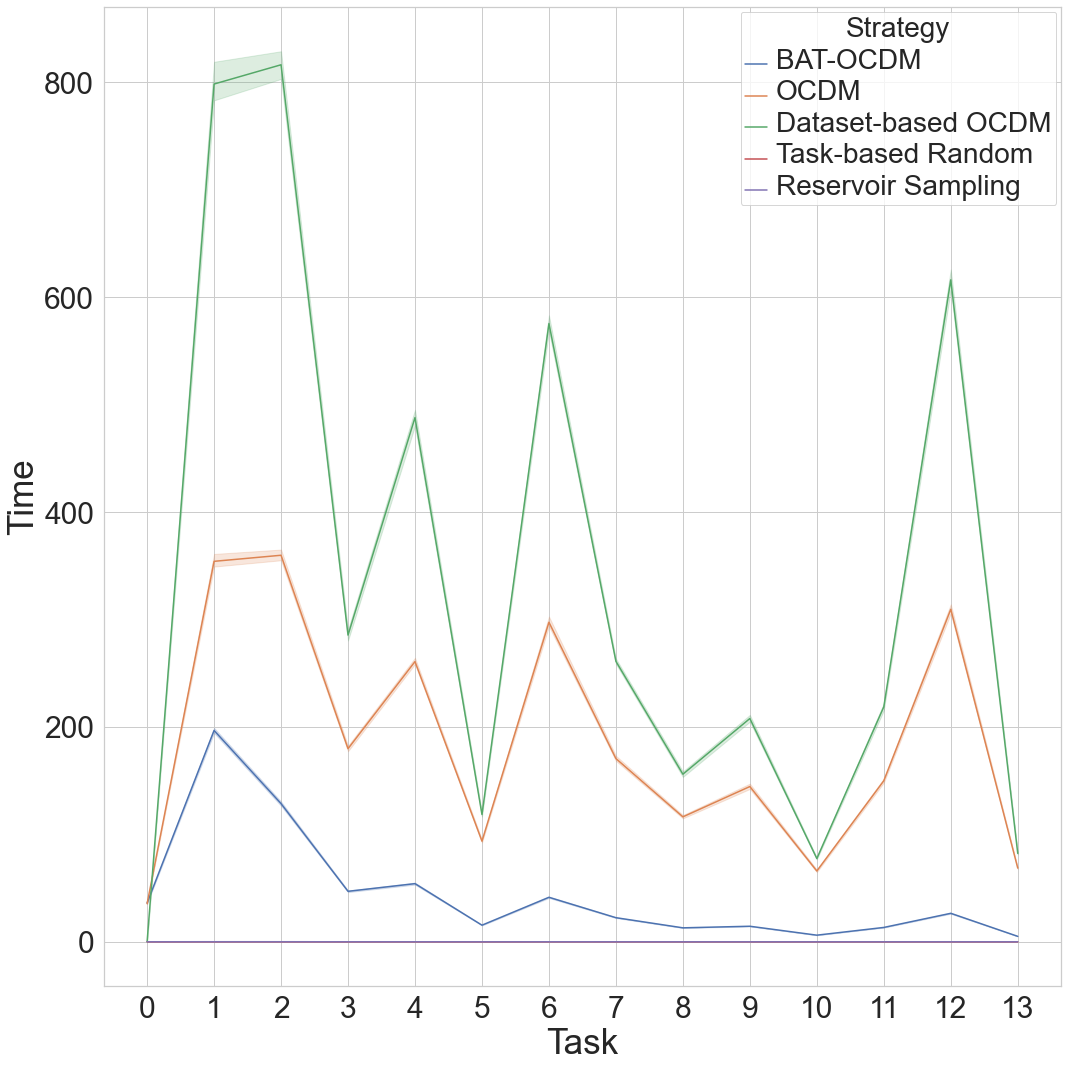}
    \caption{Computation time of each technique to handle the selection of the samples to keep in memory and remove from it. It is showed the time required during each task. On the y-axis is represented the time in seconds and on x-axis the current task.  }
    \label{fig:computational_time_plot}
\end{figure}

In Tab. \ref{tab:results_performance}, we can examine the computing time to handle the memory for the RS, OCDM, and BAT-OCDM approaches.
In comparison to traditional OCDM, BAT-OCDM requires four times less computing time to update memory. This outcome is consistent with the algorithm's anticipated theoretical complexity, which is logarithmic in the number of tasks T, as opposed to the original approach's linear in T.
Dataset-based OCDM is slower than OCDM because it takes into account all of the samples from a task at once. In practice, it is twice as slow as OCDM.\\
Additionally, the computation time (in seconds) for each task for all strategies is shown in Fig. \ref{fig:computational_time_plot}
For OCDM, we can observe that the time needed for each task is proportional to the size of the task's dataset.
In other words, the larger the task's dataset, the longer it will take.
When using dataset-based OCDM, we can observe that a task takes twice as long to complete as when using OCDM.
Finally, it appears that the BAT-OCDM computation time gradually reduces over time, which is consistent with the algorithm's logarithmic nature.

\section{Conclusions and future works}\label{sec:conclusions}
This study is the first to address multi-label classification in the context of domain incremental learning, as far as we know. For multi-label classification, earlier Continual Learning methods have focused on the Class Incremental Scenario, in which additional labels are introduced into the task sequence. Instead, in Domain Incremental Learning, the set of labels remains the same while their distribution changes over time. To overcome this problem, we present BAT-OCDM, an effective replay-based method of managing the memory update.  The proposed procedure exhibits higher performance than the simple adaptation of the previous techniques to the Domain Incremental Learning scenario, especially in the presence of class imbalance.

We validate the proposed approach on a real-world industrial Alarm Forecasting task stemming from the monitoring of packaging equipment. Experiments suggest the efficacy of the proposed methodology, especially on low-freq labels. Moreover, the complexity of BAT-OCDM is logarithmic in the number of tasks. The suggested method is therefore more effective than earlier ones that had linear complexity.
Given the efficiency of BAT-OCDM, implementation on the Edge is a viable perspective. This would be a step towards the Tiny ML paradigm, which is becoming increasingly popular, also in the scenario of the Industrial Internet of Things \citep{tinyml}.

Possible future research directions include validating BAT-OCDM performance in the Continual Incremental Learning scenario. 
Though the proposed algorithm allows for a significant reduction in the computation costs, the total complexity with regard to the dataset dimension is still quadratic. Therefore, we also envision further investigation of more efficient approaches.
Finally, an additional point to research is how to improve the selection of samples from the task. For example, the proposed approach takes into account the relation among labels while training the MLP classifier, but it does not when choosing the samples to keep in memory for the next training stage. This information, if correctly included, may lead to further performance enhancement.
\FloatBarrier

\bibliographystyle{unsrtnat}
\bibliography{biblio}

\appendix
\section{Computational Complexity}
\label{sec:appendix_computational_complexity}
Below, we are going to show the calculations that led to the results on the complexity of the algorithms present in the table of the complexity of algorithms.\\
The analysis will be performed for the original approach OCDM, the Dataset-based OCDM and our method BAT-OCDM.\\
We are going to proof that our algorithm has a complexity wrt the number of tasks T that is logarithmic, while the complexity of OCDM and Datased-based OCDM is lineart wrt T.\\
While, in the original paper is showed only the complexity for batch, we insight further and analyze the complexity for the training of an entire task and the training on all tasks.\\
We are going to analyze as first, the original approach OCDM.
In the following section, we are going to assume that each task has the same dataset size D and the memory $\mathcal{M}$ has a fixed size M.

\subsection{OCDM}

Studying the complexity for task, for each task we are going to iterate on the dataset $\mathcal{D}$ $\frac{D}{b}$ times, assuming that each batch will have size $b$.\\
In the original approach was proved that the complexity for the algorithm $Memory\_Update$(MU) using a memory of size M and removing b elements i.e. $MU(M,b)$ is:
\begin{equation}
\label{eq:complexity_memory_update}
    MU(M,b) = O( b \cdot M- \frac{b \cdot (b-1)}{2}  )
\end{equation}

OCDM algorithm use Memory\_Update to update the memory using a batch B of size $b$. In other words the algorithm MU given a set of M+b elements will remove b elements from it. The complexity of this is:

\begin{equation}
\label{eq:complexity_memory_update_batch}
    MU(M+b,b) \stackrel{(\ref{eq:complexity_memory_update})}{=} O( b \cdot (M+b)- \frac{b \cdot (b-1)}{2}  )
\end{equation}
to update a memory of size $M$ removing $b$ elements.
The complexity of algorithm OCDM for a task consist in $\frac{D}{b}$ iterations over the MU method as indicated in (\ref{eq:complexity_memory_update_batch}), which correspond to: 
\begin{equation}
\begin{split}
    OCDM_t(D,M) &= O(\frac{D}{b} \cdot [ b \cdot (M+b)- \frac{b \cdot (b-1)}{2}   ]) \\ 
    &= O( D \cdot [ M + \frac{b+1}{2} ]) \\
    &\stackrel{M>>b}{=} O(D \cdot M) 
    \label{eq:ocdm_per_task}
\end{split}
\end{equation}
Where $OCDM_t(D,M)$ is the complexity of the algorithm OCDM for a single task, assuming the number of the samples of the task as D and the memory size used as M.\\
Since the entire training correspond to iterate on T tasks, we obtain as overall performance for OCDM:
\begin{equation}
\label{eq:ocdm_totale}
    OCDM(D,M) = \sum_{t \in \mathcal{T}} OCDM_t(D,M) =  O(T \cdot D \cdot M)
\end{equation}

\subsection{Dataset-based OCDM}
Though the update per batch is essential for the Online Continual Learning(OCL) setting, we are obtain suboptimal solutions respect to find the optimal distribution using all dataset D at once.  Since we are evaluating the DIL scenario where the data D of a task is received all together we also study the performance of MU(M,D) which correspond to the of the variant \textit{Dataset-based OCDM} (Db-OCDM), while the original will remain OCDM.

\begin{align}
    Db-OCDM(D,M)_t = MU(D+M,D) \notag \\ \stackrel{(\ref{eq:complexity_memory_update})}{=} O(D \cdot (M+D) - \frac{D \cdot (D-1)}{2}) 
    \stackrel{D>>1}{=} O(M \cdot D + \frac{D^2}{2})
\end{align}

Since the entire training correspond to iterate on T tasks, we obtain as overall performance for Dataset-based:
\begin{equation}
    Db-OCDM(D,M) = O(T \cdot [ D \cdot M + \frac{D^2}{2}])
\end{equation}

\subsection{BAT-OCDM}
Below we show the complexity obtain considering our approach \textit{BAT-OCDM}.
In this case, the complexity is splitted in two subprocesses. The first one consist during the Task $i$ to select $\frac{M}{i}$ samples from the data of the new task. 
Therefore the complexity of first part is equivalent to OCDM per task i.e. $OCDM_t(D,\frac{M}{i}) $ will be $O(D \cdot \frac{M}{i})$.
\begin{equation}
\label{eq:complexity_first_part_bat_ocdm}
OCDM_t(D,\frac{M}{i}) \stackrel{(\ref{eq:ocdm_per_task})}{=} O(D \cdot \frac{M}{i})
\end{equation}

The memory of an old task must be reduced from $\frac{M}{i-1}$ to $\frac{M}{i}$, therefore eliminating $\frac{M}{i \cdot (i-1)}$ samples.\\
To do this the complete complexity is $O(\frac{M^2}{i-1})$.
In fact, we are going to perform $MU(\frac{M}{i-1}, \frac{M}{i \cdot (i-1)})$ i-1 times during task i(assuming tasks start from 1 to T included).
Therefore, for the second part we have:
\begin{equation}
    \begin{split}
    (i-1) \cdot MU(\frac{M}{i-1}, \frac{M}{i \cdot (i-1)}) &\stackrel{(\ref{eq:complexity_memory_update})  }{=}  \\
    (i-1) \cdot O( \frac{M}{i \cdot (i-1)} \cdot \frac{M}{i-1}  - \frac{M^2}{2 \cdot i^2 \cdot (i-1)^2}  ) &= \\
    O( \frac{M^2}{i\cdot(i-1)}-\frac{M^2}{2i^2\cdot(i-1)} ) &= \\
    O(\frac{M^2}{i\cdot(i-1)} \cdot ( 1-\frac{1}{2\cdot(i)} ) ) &=\\
    O( \frac{M^2}{i\cdot(i-1)} ) &= \\ 
    O(\frac{M^2}{i-1})
    \label{eq:complexity_second_part_bat_ocdm}
\end{split}
\end{equation}

Therefore, we have the total complexity for task i-th of algorithm BAT-OCDM is:
\begin{equation}
    BAT-OCDM_{t}(D,M,i) =  (\ref{eq:complexity_second_part_bat_ocdm})+(\ref{eq:complexity_first_part_bat_ocdm}) = O(\frac{D \cdot M}{i} + \frac{M^2}{i-1})
\end{equation}
Where if i=1 then the second member is 0 since during i=1 there aren't old tasks to update.
To evaluate the complexity of the entire training the calculations are trivial for OCDM and Dataset-based OCDM because it is enough to multiply the complexity by task for the value T, which is the total number of tasks.\\
In the case for BAT-OCDM is a little more tricky because the task complexity depends by task i.\\
To calculate is necessary to consider the following inequality:

\begin{equation}
    \sum_{i=1}^n \frac{1}{i} \leq \ln n +1
    \label{eq:diseq}
\end{equation}
Therefore, for the first part:
\begin{equation}
    O( \sum_{i=1}^T D \cdot M \cdot \frac{1}{i} ) \stackrel{(\ref{eq:diseq})}{=} O(D \cdot M \cdot (\ln T+1) )
    \label{eq:first_part}
\end{equation}
In the same way we have for the second part:
\begin{align}
    O( \sum_{i=2}^T \frac{M^2}{i-1} ) \stackrel{(k=i-1)}{=}
    O(\sum_{k=1}^{T-1} \frac{M^2}{k} ) \notag \\ \stackrel{(\ref{eq:diseq})}{=} O( M^2 \cdot [\ln (T-1) + 1]) 
    = O( M^2 \cdot [\ln (T) + 1] )
    \label{eq:second_part}
\end{align}
In total we have:
\begin{align}
    BAT-OCDM(D,M) = (\ref{eq:first_part})+(\ref{eq:second_part}) \notag \\
    = O( (\ln T +1) \cdot (D \cdot M + M^2) )
\end{align}


\end{document}